# Noise Adaption Network for Morse Code Image Classification

Xiaxia Wang, XueSong Leng, and Guoping Xu[*]

Hubei Key Laboratory of Intelligent Robot, School of Computer Sciences and Engineering, Wuhan Institute of Technology, Wuhan, Hubei, China
`xugp@wit.edu.cn`

**Abstract.** The escalating significance of information security has underscored the pervasive role of encryption technology in safeguarding communication content. Morse code, a well-established and effective encryption method, has found widespread application in telegraph communication and various domains. However, the transmission of Morse code images faces challenges due to diverse noises and distortions, thereby hindering comprehensive classification outcomes. Existing methodologies predominantly concentrate on categorizing Morse code images affected by a single type of noise, neglecting the multitude of scenarios that noise pollution can generate. To overcome this limitation, we propose a novel two-stage approach, termed the Noise Adaptation Network (NANet), for Morse code image classification. Our method involves exclusive training on pristine images while adapting to noisy ones through the extraction of critical information unaffected by noise. In the initial stage, we introduce a U-shaped network structure designed to learn representative features and denoise images. Subsequently, the second stage employs a deep convolutional neural network for classification. By leveraging the denoising module from the first stage, our approach achieves enhanced accuracy and robustness in the subsequent classification phase. We conducted an evaluation of our approach on a diverse dataset, encompassing Gaussian, salt-and-pepper, and uniform noise variations. The results convincingly demonstrate the superiority of our methodology over existing approaches. The datasets are available on https://github.com/apple1986/MorseCodeImageClassify

## 1      Introduction

Morse code, widely used in telecom, employs short and long signals, applied in wireless broadcasting, IoT, positioning, and remote control [1][2]. However, noise during transmission often compromises Morse code image accuracy, leading to reduced classification precision. Current strategies, employing manual feature extraction and traditional machine learning for Morse code image classification, are time-consuming and perform suboptimally in the presence of noise [4]. Manual Morse code deciphering is labor-intensive and less stable, necessitating the development of automatic recognition methods such as AUSP[5], LSM[6], and AVRTP[7].

To tackle these challenges, adaptive methodologies[8][9] aim to improve Morse code recognition precision by addressing inherent uncertainty and reducing noise pollution. Some also attempt noise reduction through preprocessing techniques[10]-[15]. However, these methods often entail time-consuming preprocessing and complex identification procedures, adding to their complexity.

In recent years, deep learning (DL) [3] has demonstrated remarkable capabilities in image classification, eliminating the need for manual feature extraction. While DL-powered end-to-end solutions for Morse Code Image Classification Tasks have exhibited exceptional performance [16][17], they are often tailored for specific noise issues and may not effectively address scenarios with diverse noise sources.

In response, our work introduces a deep learning approach to enhance Morse code image classification accuracy amidst diverse noise types. Our methodology unfolds in two stages: firstly, a U-Net-like sub-network is designed for noise adaptation, capturing discriminative representations and eliminating noise through decoding. Secondly, a deep convolutional neural network with increasing convolution kernels is developed to acquire intricate features, improving precision and adaptability in Morse code classification. In summary, this paper contributes:

1) A Multi-noise Morse code image dataset, including pristine images and three noise-contaminated categories, offering a thorough evaluation of noise interference in Morse image classification.

2) An innovative two-stage framework integrating denoising and classification through a noise-adaptive approach for Morse code image recognition.

3) Pioneering classification of Morse code images under various noise types, backed by extensive experiments verifying its effectiveness.

## 2      Related Work

The autoencoder, an unsupervised learning algorithm, focuses on compressing and reconstructing data to improve feature representation, reconstruction accuracy, and data compression [18]. Addressing image noise, Huang [19] utilized denoising filters, while Jaspin-Jeba-Sheela [20] introduced the median filter (MF) for edge preservation. Zhang [21] proposed a wavelet-based filter for reducing additive noise. However, manual parameter tuning and inherent limitations are common with these methods. To tackle these challenges, Pascal Vincent [22] introduced autoencoder denoising, enhancing feature representation and characterizing data by learning restoration from noisy versions.

## 3      Methodology

This paper introduces "Noise Adaptation Network" (NANet) , a novel approach for Morse code image classification, comprising two stages: an initial stage with an Adaptive AutoEncoder to reduce noise and a subsequent stage using expanding con-volution kernels to enhance classification accuracy and generalization(see 错误!未找到引用源。).

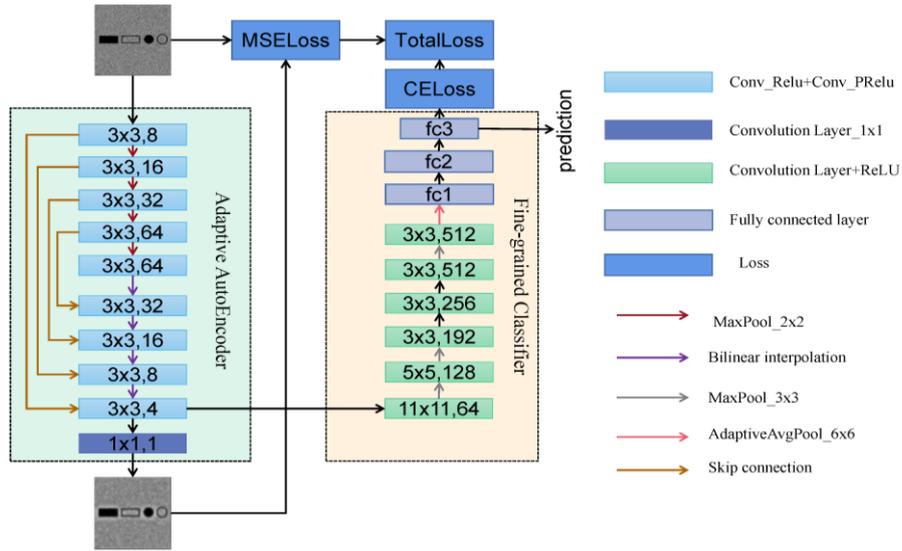

**Fig. 1.** The NANet for Morse code image classification features a UNet-like subnetwork comprising an encoder, decoder, and classifier.

### 3.1 Adaptive AutoEncoder

The Adaptive Autoencoder enhances feature representation and spatial detail capture through distinct encoding and decoding phases, aiming to improve the network's ability to identify key features and understand spatial complexities. The encoding phase gradually extracts advanced features using convolutional and pooling layers, employing a dual-convolution approach to prevent overfitting and enhance generalization. Skip connections aid in information flow and feature fusion, while max pooling during downsampling preserves details and reduces noise impact. Transposed convolutions during upsampling enhance detail clarity, while a final convolutional layer maintains image characteristics, producing high-fidelity denoised images.

### 3.2 Fine-grained Classifier

The nine-layer fine-grained classifier is pivotal in the architecture, with initial convolutional operations, ReLU activations, and max pooling in the first six layers. Three fully connected layers follow, enhancing comprehension and classification on denoised images. Incorporating three distinct convolutional kernel sizes (11x11, 5x5, and four 3x3) improves adaptability to diverse image variations. Adaptive average pooling before the fully connected layers boosts generalization and prepares features for further processing. Each fully connected layer has 2048 neurons, transforming convolutional features into class scores. Integrated dropout layers effectively combat overfitting, promoting better generalization and performance stability.

### 3.3 Loss Function

The model uses mixed error training, tackling reconstruction error and classification loss together. An adaptive autoencoder reconstructs input data to minimize differences with pristine data using the mean squared error (MSELoss) function, prioritizing larger errors during optimization for improved accuracy and fidelity. Mathematically, this is shown as:

$$MSELoss = \frac{1}{n}\sum_{i=1}^{n}(\hat{y}_i - y_i)^2 \quad (1)$$

where $y_i$ represents the true data, $\hat{y}_i$ represents the predicted data, and $n$ represents the total number of samples.

The cross-entropy loss function was used for classification in this study, which is calculated as follows:

$$CELoss = -\frac{1}{n}\sum_{i=1}^{n}\sum_{j=1}^{C} y_{i,j} \log(\hat{y}_{i,j}) \quad (2)$$

where $n$ represents the number of samples, $C$ represents the number of classes, $y_{i,j}$ represents the true label of the $i$-th sample in the $j$-th class, and $\hat{y}_{i,j}$ represents the predicted probability of the $i$-th sample in the $j$-th class. The training in this study used the sum of the reconstruction error and the classification loss, which is calculated as follows:

$$TotalLoss = MSELoss + CELoss \quad (3)$$

## 4 Experimental Result

### 4.1 Experimental details

Experiments used Python 3.7.15 and PyTorch 1.9.1, with random rotation for dataset diversification. Optimization employed the Adam optimizer with a fixed learning rate of 0.0001, while input images were standardized to 224x224 for computational efficiency on Nvidia 1080 GPU. Training comprised batches of eight over 300 epochs, with model evaluation based on accuracy, precision, recall, and F1 score for a comprehensive assessment of classification efficacy.

$$Accuracy = \frac{(TP+TN)}{(TP+TN+FP+FN)} \quad (4)$$

$$Precision = \frac{TP}{(TP+FP)} \quad (5)$$

$$Recall = \frac{TP}{(TP+FN)} \quad (6)$$

$$F1 = 2 * \frac{(Precision * Recall)}{(Precision + Recall)} \tag{7}$$

where *TP* represents True Positive, *TN* represents True Negative, *FP* represents False Positive, and *FN* represents False Negative. The final results are averaged for *Precision*, *Recall*, and *F1* scores for each class.

### 4.2 Dataset

This study centers on Morse code image classification, using rectangles and circles for long dashes and dots. Our dataset comprises 1040 images, with 40 pristine Morse code images for each of the 26 English letters, all at a uniform resolution of 512x512 pixels. To gauge the proposed approach's robustness, we assessed its performance under three types of noise: uniform, Gaussian, and salt-and-pepper. Fig. 2 illustrates four instances of Morse code images depicting the letter 'A,' showcasing the dataset's visual diversity.

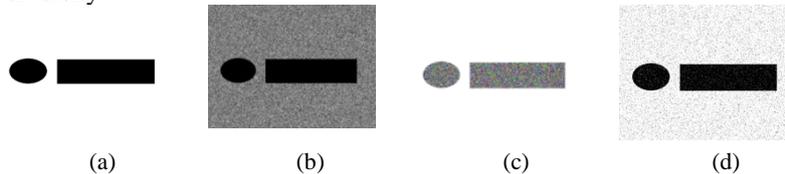

(a)  (b)  (c)  (d)

**Fig. 2.** Examples of images in the clean dataset (a), uniform noise dataset (b), Gaussian noise dataset (c), and salt-and-pepper noise dataset (d) corresponding to the letter A.

### 4.3 Performance Comparison with State-of-the-Art MethodsDataset

We ensured impartiality by consistently applying settings to classification networks and our NANet, training on a clean Morse code dataset and testing on four image types, each with 10 images. As demonstrated in Tables 1 to 3, our model excels in classifying clean datasets and exhibits remarkable resilience to various types of noise, outperforming established networks in classifying noisy images. The network's CAM visualization (see Fig.3). underscores the significance of focused regions in image classification, independent of input image noise levels.

**Table 1.** Test results of models on Uniform Noise Morse Code dataset

| Method | Accuracy(%) | Precision (%) | Recall (%) | F1 Score (%) |
|---|---|---|---|---|
| MobileNetV2 | 7.31 | 12.71 | 7.31 | 4.29 |
| SqueezeNet | 8.85 | 4.08 | 8.85 | 4.38 |
| Shufflenetv2 | 4.23 | 4.61 | 4.23 | 1.64 |
| AlexNet | 71.92 | 80.90 | 71.92 | 71.37 |
| VGG11 | 84.62 | 88.95 | 84.62 | 84.88 |
| VGG13 | 87.31 | 89.96 | 87.31 | 87.37 |
| VGG16 | 79.23 | 82.97 | 79.23 | 79.21 |
| NANet | **96.92** | **97.16** | **96.92** | **96.93** |

**Table 2.** Test results of models on Gaussian Noise Morse Code dataset

| Method | Accuracy(%) | Precision (%) | Recall (%) | F1 Score (%) |
| --- | --- | --- | --- | --- |
| MobileNetV2 | 62.69 | 74.16 | 62.69 | 60.74 |
| SqueezeNet | 80.38 | 83.85 | 80.38 | 79.97 |
| Shufflenetv2 | 31.15 | 49.76 | 31.15 | 29.36 |
| AlexNet | 76.92 | 80.31 | 70.92 | 76.69 |
| VGG11 | 92.69 | 93.38 | 92.69 | 92.71 |
| VGG13 | 92.31 | 92.93 | 92.31 | 92.16 |
| VGG16 | 90.38 | 91.12 | 90.38 | 90.37 |
| NANet | **93.46** | **94.66** | **93.46** | **93.46** |

**Table 3.** Test results of models on Salt-and-Pepper Noise Morse Code dataset

| Method | Accuracy(%) | Precision (%) | Recall (%) | F1 Score (%) |
| --- | --- | --- | --- | --- |
| MobileNetV2 | 5.77 | 8.81 | 5.77 | 3.29 |
| SqueezeNet | 15.00 | 12.93 | 15.00 | 8.01 |
| Shufflenetv2 | 8.85 | 3.64 | 8.85 | 3.35 |
| AlexNet | 83.08 | 86.80 | 83.08 | 83.03 |
| VGG11 | 95.00 | 95.37 | 95.00 | 94.94 |
| VGG13 | 92.69 | 93.27 | 92.69 | 92.53 |
| VGG16 | 85.38 | 86.71 | 85.38 | 85.14 |
| NANet | **98.46** | **98.57** | **98.46** | **98.46** |

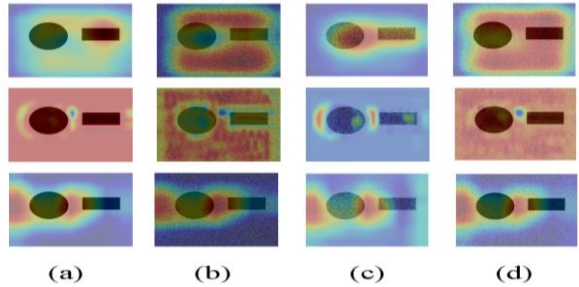

**Fig. 3 .** CAM visualization of VGG16 and our model on four datasets: clean, uniform noise, Gaussian noise, and salt and pepper noise datasets, presented top to bottom.

### 4.4 Ablation study

To validate our adaptive autoencoder's effectiveness, we conducted ablation experiments on five representative models. Using denoised feature maps generated by the adaptive autoencoder as inputs, we trained the five baseline models and tested them on four datasets. Results in Table 4 demonstrate a general improvement in testing accuracy for various noisy datasets after applying our adaptive autoencoder for denoising. This underscores the autoencoder's efficacy in adapting to noise and improving generalization in other classification models, as discussed here.

Table 4. The test accuracy (%) results of models denoised with an automatic denoising encoder on four datasets, comparing training with the benchmark model alone (left) versus denoising with the encoder before training (right). Salt represents the Salt-and-Pepper dataset.

| Method | Clean (%) | | Uniform (%) | | Gaussian(%) | | Salt(%) | |
|---|---|---|---|---|---|---|---|---|
| MobileNetV2 | 97.31 | 91.54 | 7.31 | 10.00 | 62.69 | 79.62 | 5.77 | 15.38 |
| SqueezeNet | 98.85 | 98.46 | 8.85 | 6.92 | 80.38 | 88.85 | 15.00 | 26.15 |
| AlexNet | 92.31 | 97.31 | 71.92 | 86.92 | 76.92 | 95.00 | 83.08 | 95.77 |
| VGG16 | 92.31 | 95.38 | 79.23 | 69.62 | 90.38 | 93.08 | 85.38 | 91.54 |
| NANet | 98.46 | | 96.92 | | 93.46 | | 98.46 | |

## 5 Conclusion

This paper proposes a two-stage method for classifying noisy Morse code images, involving a UNet-like sub-network for denoising in the first stage and a deep convolutional neural network for classification in the second stage. Compared to state-of-the-art methods like VGG and MobileNet, NANet demonstrates superior classification accuracy and robustness across three types of noise-polluted Morse code images: uniform noise, Gaussian noise, and salt-and-pepper noise. Future research will explore the effectiveness of our method in other computer vision tasks such as object detection and instance segmentation.

## Acknowledgement

This work is supported by the Guangdong Provincial Key Laboratory of Human Digital Twin (No. 2022B1212010004), the Fundamental Research Funds for the Central Universities of China (No. PA2023IISL0095), and the Hubei Key Laboratory of Intelligent Robot in Wuhan Institute of Technology (No. HBIRL 202202).